# 基于条件深度卷积生成对抗网络的视网膜血管分割

蒋芸；谭宁

(西北师范大学 计算机科学与工程学院,兰州 730000)

**摘 要** 视网膜血管的分割帮助医生对眼底疾病进行诊断有着重要的意义。但现有方法对视网膜血管的分割存在着各种问题,例如对血管分割不足,抗噪声干扰能力弱,对病灶敏感等。针对现有血管分割方法的缺陷,本文提出使用条件深度卷积生成对抗网络的方法对视网膜血管进行分割。我们主要对生成器的网络结构进行了改进,在卷积层引入残差模块进行差值学习使得网络结构对输出的改变变得敏感,从而更好的对生成器的权重进行调整。为了降低参数数目和计算,在使用大卷积核之前使用小卷积对输入特征图的通道数进行减半处理。通过使用跳远连接将卷积层的输出与反卷积层的输出进行连接从而避免低级信息共享。通过在DRIVE和STARE数据集上对本文的方法进行了验证,其分割准确率分别为96.08%、97.71%,敏感度分别达到了82.74%、85.34%,F-measure分别达到了82.08%和85.02%,灵敏度比R2U-Net的灵敏度分别高了4.82%,2.4%。

**关键词** 生成对抗网络,残差网络,视网膜血管分割,条件模型,卷积神经网络

## Retinal Vessel Segmentation Based on Conditional Deep Convolutional Generative Adversarial Networks

JIANG Yun, TAN Ning

(School of Computer Science and Engineering, Northwest Normal University,Lanzhou 730000,China)

**Abstract** The segmentation of retinal vessels is of significance for doctors to diagnose the fundus diseases. However, existing methods have various problems in the segmentation of the retinal vessels, such as insufficient segmentation of retinal vessels, weak anti-noise interference ability, and sensitivity to lesions, etc. Aiming to the shortcomings of existed methods, this paper proposes the use of conditional deep convolutional generative adversarial networks to segment the retinal vessels. We mainly improve the network structure of the generator. The introduction of the residual module at the convolutional layer for residual learning makes the network structure sensitive to changes in the output, as to better adjust the weight of the generator. In order to reduce the number of parameters and calculations, using a small convolution to halve the number of channels in the input signature before using a large convolution kernel. By used skip connection to connect the output of the convolutional layer with the output of the deconvolution layer to avoid low-level information sharing. By verifying the method on the DRIVE and STARE datasets, the segmentation accuracy rate is 96.08% and 97.71%, the sensitivity reaches 82.74% and 85.34% respectively, and the F-measure reaches 82.08% and 85.02% respectively. The sensitivity is 4.82% and 2.4% higher than that of R2U-Net.

## 1 引言

血管作为血液循环流动的管道,遍布人体的各个部位,尤其是眼底中含有大量的动脉血管,是人身体内唯独采用无损伤手段就能够直接观察到的较深层次的微血管系统,为医生提供了有关眼睛状况和一般系统状态的丰富信息。 眼科医生可以检测到高血压和糖尿病引起的全身血管负荷增加以及视网膜血管疾病如视网膜静脉阻塞和视网膜动脉阻塞等血管结构异常的早期征兆,血管和血管系统引起的疾病会造成患者失明。随着技术的发展,探索一种自动化方法分割视网膜血管已经被广泛研究。从而辅助医师诊断、分析患者病情,对眼底疾病做出尽早的预防和诊治,可有效避免绝病变所引发的视觉损失。

目前在国内外的血管分割算法有匹配滤波器法[1]、多阈值的血管检测[2]、以形态学为基础技

术血管分割[3]、区域生长法、使用神经网络的血管分割算法[4-6]、多尺度层分解和局部自适应阈值血管分割方法[7]、基于活动轮廓模型的血管分割[8]以及基于模糊聚类的分割方法[9]等。文献[10]提出了一种基于多尺度 2D Gabor 小波变换和形态学重构的血管分割方法。采用不同尺度的 2D Gabor 小波对视网膜图像进行变换，并分别应用形态学重构和区域生长法对变换后的图像进行分割，最后，对以上两种方法分割的视网膜血管和背景像素点重新标记识别，得到视网膜血管最终分割结果，但对于血管与背景对比度低的图片，分割假阳性率高。文献[11]提出一种融合区域能量拟合信息和形状先验的水平集血管分割方法，通过形态学算子去除血管中心亮线，并与原图像和掩模分别进行减法和点乘运算增强视网膜血管图像，然后分析 Hessian 矩阵的特征值在血管、背景和病灶上不同的几何性质，利用 Hessian 矩阵特征值重构血管响应函数，最大化不同结构的差异，从而获得视网膜血管初步图像，但因为高斯卷积算子的影响，导致复杂的血管交叉处分割不足。文献[12]提出通过三阶段对血管进行分割，首先对眼底图像的绿色平面进行预处理并针对血管区域中提取二值图像，然后使用基于像素邻域一阶和二阶梯度图像提取的八个特征集合，使用高斯混合模型（GMM）分类器对两个二值图像中的所有其余像素进行分类，血管的主要部分与分类的血管像素结合，但存在微血管易断的问题。文献[13]使用将分割任务重塑为从视网膜图像到血管图的跨模态数据转换的分割监督方法，提出了一种具有较强诱导能力的广泛深度神经网络模型，并给出了一种有效的训练策略。文献[14]一种将异构情境感知功能与区分性学习框架相结合的方式对眼底血管图片进行分割。文献[15]使用监督分割技术，该技术使用在全局对比度归一化，零相位白化以及使用几何变换$\gamma$和校正进行预处理的样本上训练深度神经网络，对噪声有较强的适应能力，但依旧无法克服微血管易断的问题，分割的准确率也有待提高。综上所述，虽然国内外研究人员已提出很多血管分割方法，但大部分分割的结果精度不高，性能也有很大的改善空间。针对现有方法对视网膜血管分割不足，准确率不高对噪声、病灶敏感等问题，于是使用生成式思想分割出高精度，高准确率的视网膜血管图像。本文主要工作包括：

1、对于之前所提出的方法中存在的局限性进行了详细的分析后，使用生成式框架针对视网膜血管分割效果不理想的问题，实现了一种自动分割视网膜血管的方法。

2、提出了一个新的网络结构如图3，利用残差网络的思想，使得网络结构对输出的变化和权重的改变变得敏感，从而更好的对权重进行了调整，提高分割的效果，也缓解了梯度减少，解决了梯度消失的问题。在使用3×3的卷积之前使用1×1的卷积作为瓶颈层对输入的特征图进行降维，然后用3×3的卷积核对维度进行还原，降低了网络的复杂度，保持了精度又减少了计算量。

3、由于编码器-解码器(Encoder-Decoder)这种网络结构要求所有的信息流通过所有的网络层，在输入和输出之间共享大量本可直接穿过网络层的低级信息。为了避免出现这种情况，本文使用了跳远连接(skip connection)，对于$n$层的网络，每个跳远连接将第$i$层网络的输出和第$n-i$层的输出进行连接，作为第$n-i+1$层的输入。

## 2 生成对抗

近些年来，无监督学习已经成为了研究的热点，变分自编码器[16]、生成对抗网络(Generative Adversarial Nets，GAN)[17]等无监督模型受到越来越多的关注。在人工智能高速发展的时代，GAN 的提出不仅满足了相关领域的研究和应用和需求，也带来了新的发展动力。特别是在图像和视觉领域中对 GAN 的研究和应用是最为广泛，

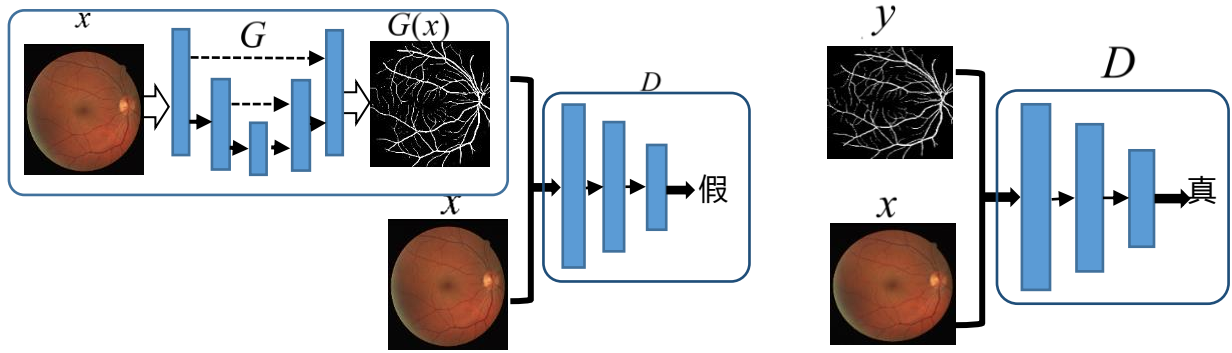

图1 视网膜血管图像分割模型

Fig. 1 Retinal vessels image segmentation model

已经可以通过随机数字生成人脸、从低分辨率图像生成高分辨率图像等。此外，GAN 已经开始逐渐应用到医学图像处理中：模拟超声探头的空间位置上有条件地对解剖学精确的图像进行采样[19]、检测恶性前列腺癌[20]等问题的研究中。

### 2.1 生成对抗网络原理

生成对抗网络由生成器 G 和判别器 D 两部分组成，生成器主要负责通过随机噪声生成服从真实数据分布的样本，判别器负责通过识别输入的数据样本来自生成器还是真实数据，两者相互迭代优化提升各自的性能，最终使得判别器无法判别输入的数据来源时，则认真生成器以学得了真实数据的分布。其损失函数为：

$$\max_D \min_G (\theta_G, \theta_D) = \mathbb{E}_{x \sim pdata}[\log D(x)] + \mathbb{E}_{z \sim pz(z)}[\log(1 - D(G(z)))] \quad (1)$$

### 2.2 条件生成对抗网络

条件生成对抗网络(CGAN)是在 GAN 的基础上加入了辅助信息 y，通过额外的辅助信息 y 控制生成器对数据的生成。y 可以是任何类型的辅助信息，比如类标签。其模型结构图 2，

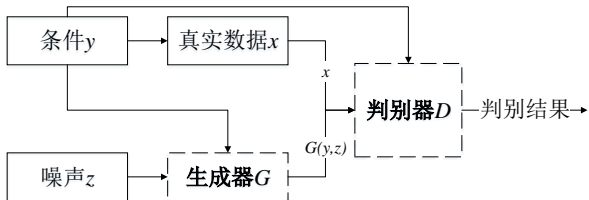

图2 条件生成对抗网络模型

Fig. 2 Condition generation against network model

损失函数表示为：

$$\min_G \max_D \mathcal{L}(D,G) = \mathbb{E}_{x \sim pdata(x)}[\log D(x|y)] + \mathbb{E}_{z \sim pz(z)}[\log(1 - D(G(x|y)))] \quad (2)$$

## 3 基于条件深度卷积生成对抗网络的视网膜血管分割

本文通过结合深度卷积生成对抗网络(Deep Convolutional Generative Adversarial Networks，DCGAN)[21]和U形卷积神经网络[22]的特点，在生成器使用U型卷积神经网络模型的思想，利用U型卷积网络的边缘检测能力对视网膜图像中的血管进行分割，将分割出的视网膜血管图像和原图像一同输入判别器中进行判断，最后直到判别器无法正确区分输入的视网膜血管图像的来源（专家分割或者生成器分割）。

生成对抗神经网络(GAN)的学习优化过程是寻找到生成器和判别器之间的一个纳什均衡，生成对抗网络分割眼底视网膜问题的目标函数如下所示：

$$\mathcal{L}_{cGAN}(G,D) = \mathbb{E}_{x \sim pdata(x), y}[\log D(x,y)] + \mathbb{E}_{x \sim pdata(x)}[\log(1 - D(x, G(x)))] \quad (3)$$

$x$ 为输入的视网膜图像，$y$ 为专家分割的视网膜血管图像，$D(x,y)$ 表示视网膜血管图像来源于专家分割的概率，$D(x,G(x))$ 表示视网膜血管图像来自生成器分割的概率。生成器 ($G$) 尝试最

小化目标函数，判别器($D$)尝试最大化目标函数，$GAN$通过对下面的函数进行优化：

$$G^* = \min_G \max_D \mathcal{L}_{cGAN}(G, D) \quad (4)$$

通过实验发现，将现有的$GAN$目标函数与传统的损失（例如$L1$距离函数）相结合时，生成出的视网膜血管图像更加趋近于专家分割的视网膜血管图像，判别器的作用保持不变，但生成器的任务不仅要欺骗判别器，而且要最小化生成的视网膜血管图像与人工分割出的视网膜血管图像直接的$L1$距离。

$$\mathcal{L}_{L1}(G) = E_{x \sim pdata(x), y}[||\, y - G(x, y)\,||] \quad (5)$$

通过将$GAN$的目标函数与$L1$距离函数进行结合，最终的目标函数如下：

$$G^* = \arg\min_G \max_D \mathcal{L}_{cGAN}(G, D) + \lambda \mathcal{L}_{L1}(G) \quad (6)$$

$\lambda$用于平衡两个目标函数，推荐$\lambda$的值为100

## 4 模型结构

分割视网膜血管的模型如图1所示，通过输入一张视网膜图像$x$到生成器($G$)中，生成器通过学习到从眼底视网膜图像$x$到眼底血管图像$y$之间的映射关系，$G: x \to y$，最终输出一张分割后的视网膜血管图像。判别器($D$)通过学习输入图像对$\{x, y\}$与$\{G(x), y\}$之间的分布差异从而正确判别输入图像对来源的二分类器$\{0,1\}$，判别器输出1表示输入的视网膜血管图像来源于人工分割，输出0表示视网膜血管图像来源于生成器。

### 4.1 生成器网络结构

生成器中使用$x \in R^{w \times h \times c}$作为输入图像，$w = n = 512$，$c = 3$，网络结构如图3所示，之前许多分割方法的网络结构使用编码器-解码器(Encoder-Decoder)[23]，这种网络结构通过向下采样，逐渐降低采样层，直到达到一个瓶颈层，将提取的信息变为一个一维向量，在这一点上进行反卷积，逐渐向上采样，最后还原成图像。这样的网络结构要求所有的信息流通过所有的网络层，包括瓶颈层。在许多图像分割问题中，输入和输出之间共享大量本可直接穿过网络层的低级信息。为了使生成器能够避免出现这种情况，本文使用了跳远连接(skip connection)，对于$n$层的网络结构，每个跳远连接将第$i$层网络的输出和第$n-i$层的输出进行连接，作为第$n-i+1$层节点的输入。 在编码层使用$LReLU$作为非线性激活函数，每层都是用批量归一化(Batch-Normalization, BN)[24]。通过归一化当前输入($\mu = 0, \sigma = 1$)，有利于加速整个网络的学习，提高卷积层之间的独立性。解码层使用$ReLU$作为非线性函数，在最后一层使用$Tanh$激活函数生成图片。为了提高模型分割的准确率，我们在生成器中加入残差网络结构，结构如图4(a)所示，由于残差可以缓解梯度减少，解决梯度消失的问题，提高对网络结构种权重改变的敏感度，使得生成器能够充分学习到视网膜血管图像的分布，从而提高分割的效果。 为了降低模型的复杂度，减少计算量和训练的参数数量，在每次使用3×3的卷积核之前加入1×1卷积核作为瓶颈层对输入的特征层进行降维。将通道数目降为原

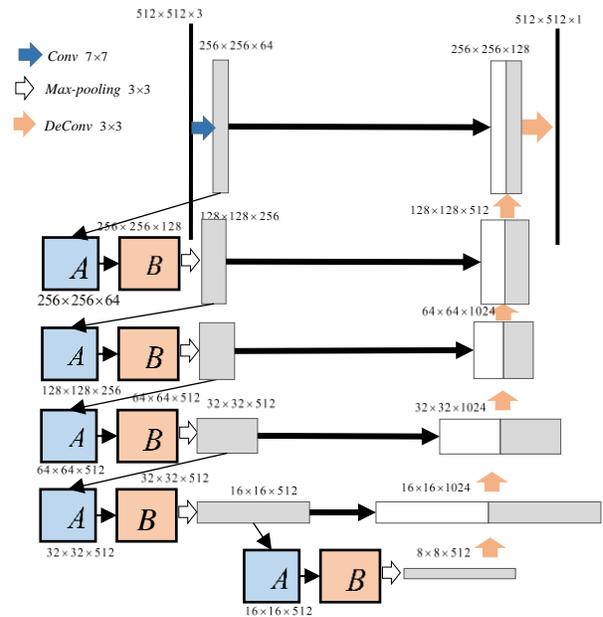

图3 生成器网络结构
Fig. 3 Discriminator network structure

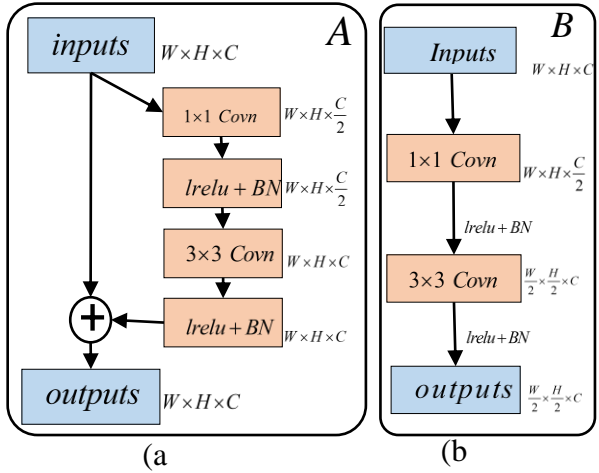

图 4 卷积的不同变体 (a)残差卷积单元 (b)前向传播卷积单元

Fig. 4 Different variant of convolutional (a) Forward convolutional units, (b) Residual convolutional unit

来的一半，然后通过 3×3 卷积核恢复到原来的通道数，具体结构如图 4(b)所示。

### 4.2 判别器网络结构

判别器的网络结构如图 5 所示，使用 $\{x_i, y_i\}_{i=1}^N$ 或者 $\{x_i, G(x_i)\}_{i=1}^N$（$x_i \in X, y_i \in Y$）作为输入，$x \in R^{512 \times 512 \times 3}$、$y \in R^{512 \times 512 \times 1}$。对于输入为 $\{x_i, y_i\}$ 时，判别器正确的输出为 1，输入 $\{x_i, G(x_i)\}$ 判别器正确的输出为 0。在判别器中的每一个卷积层中使用 LeakyReLU 作为非线性激活函数，使用带步幅(Strided convolutions)的卷积代替池化层(pooling)的效果。

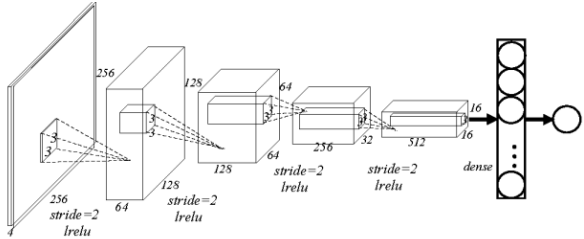

图 5 判别器网络结构

Fig. 5 Discriminator network structure

## 5 训练

本论文工作基于深度卷积对抗神经网络(DCGAN)和条件生成对抗神经网络(cGAN)[25]实现，使用 DCGAN 中推荐的训练参数进行训练，训练时使用 Adam 方式进行梯度下降（$\beta_2=0.999$，$\beta_1=0.5$，$\varepsilon=10^{-8}$），学习率为 $lr=0.0002$，$mini-batch=1$，每层输出的结果进行批量归一化从而减少每层之间的依赖性，从而提高各网络层之间的独立性，卷积时使用 LeakyReLU 激活函数，leak 斜率设置为 0.2，反卷积时，使用 ReLU 激活函数。迭代 800 个周期，每次输入是随机对视网膜图像进行数据增强操作，输出的分割图像的大小为 $512 \times 512 \times 1$

生成器的训练过程如图 6 所示，主要通过两个途径调整权重：（1）通过比较生成器分割出的视网膜图像 $G(x)$ 与专家分割出的血管图像 $y$ 之间的差值，通过差值调整权重，使 $\|y-G(x)\|$ 的值最小化；（2）将视网膜图像 $x$ 与生成器分割出的血管图像 $G(x)$ 输入到判别器中，根据判别器输出的结果 $D(G(x), y)$ 与 1 之间的差值对生成器的权重进行调整。通过上面两种方式对生成器权值的调整，使生成器分割出的视网膜血管图像更加接近专家分割后的血管图像。

判别器的训练过程如图 7 所示：也是通过两个途径调整权重：（1）输入视网膜图像 $x$ 和专家分割的视网膜血管图像 $y$，通过比较判别器输出的值 $D(x, y)$ 与标准值 1 直接差值调整权重，使判别器输出的值更加接近 1；（2）输入视网膜图像 $x$ 和生成器分割出的血管图像 $G(x)$，比较判别器输出的结果 $D(G(x), y)$ 与 0 之间的差值，通过差值调整权重。通过这两种方式对权值的调整，使得判别器能更准确的区分视网膜血管图像是通过专家分割的还是来自生成器分割。

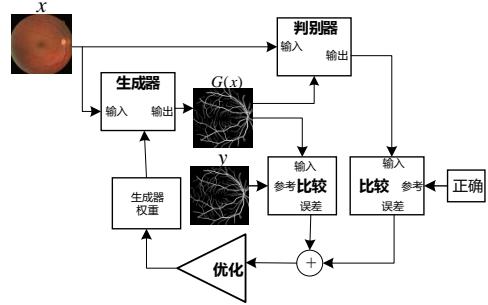

图 6 生成器训练的过程

Fig. 6 Generative training process

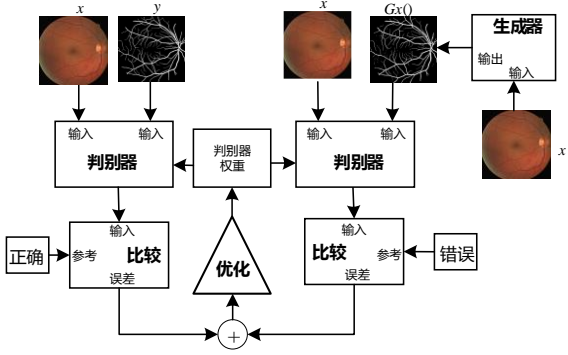

图 7 判别器训练的过程

Fig. 7 Discriminator training process

## 6 实验结果与分析

我们在两个公开的眼底视网膜数据集（*DRIVE*[26]和 *STARE*[10]）上对我们的方法进行了测试，在 *DRIVE* 数据集中，总共包含 40 张视网膜图像和对应的视网膜血管图像，用前 20 张图像做训练样本集，后 20 张图像做测试样本集，我们在第一类视网膜血管标签图像上进行训练和测试。在 *STARE* 数据集中我们用前 10 张图像做训练样本集，后 10 张图像当测试集。由于数据集中只有少量的样本集对网络结构进行训练，但深度神经网络对参数的训练需要大量的训练样本，这造成了数据的饥饿性，我们通过对图像的旋转、水平翻转、垂直翻转、平移变换、图像灰度值的变化等方法对训练数据集进行扩充，这对于提高分割的准确率、防止过拟合、和网络的鲁棒性至关重要。

### 6.1 不同分割算法结果的对比

我们分别在 STARE、DRIVE 数据集上进行了大量的血管分割实验来证明方法的可行性，图 8 和图 9 分别展示了在 DRIVE 数据集上和 STARE 数据集上文献[31]、文献[36]与本文方法对其中两张视网膜图像的分割结果的对比，在图 8 和图 9 中，(a)为原始视网膜图像，(b)为专家分割的视网膜血管图像，(c)和(d)分别为文献[36]和文献[31]的分割结果，其中图 9(a)中第一张和第二张分别为健康人，青光眼视网膜图像，尽管图 9 (a)的第二张图像中有血管有病变，存在病灶，但本文的方法可以较好的克服这些因素，分割出一张准确的血管图像，从图 8 和图 9 的(c)和 (d)中可以看出文献[36]和文献[31]分割的视网膜血管图像中包含了大量的噪声，对病变视网膜图像

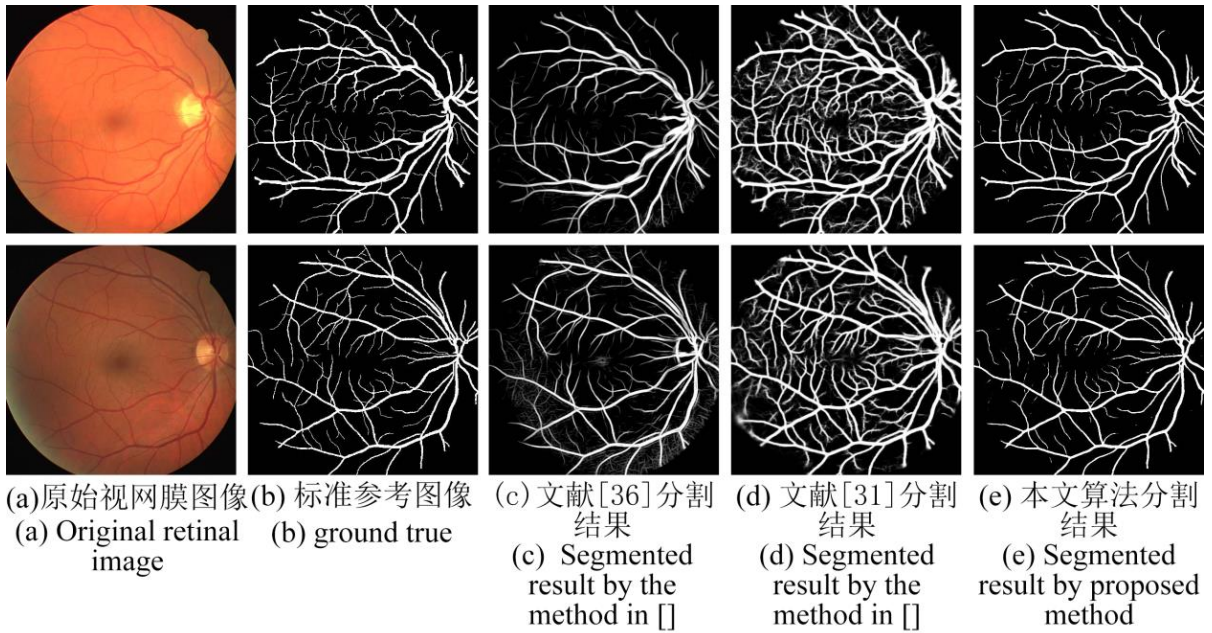

(a)原始视网膜图像 (b) 标准参考图像 (c) 文献[36]分割结果 (d) 文献[31]分割结果 (e) 本文算法分割结果

(a) Original retinal image (b) ground true (c) Segmented result by the method in [] (d) Segmented result by the method in [] (e) Segmented result by proposed method

图 8 DRIVE 数据库视网膜血管分割结果比较

Fig. 8 Comparisons of segmentation results on DRIVE database

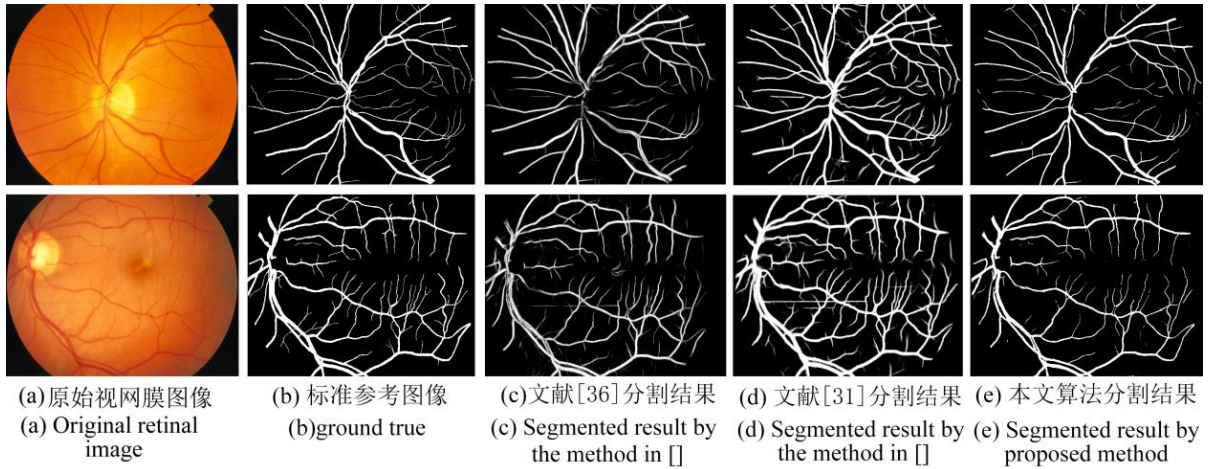

(a) 原始视网膜图像　(b) 标准参考图像　(c) 文献[36]分割结果　(d) 文献[31]分割结果　(e) 本文算法分割结果
(a) Original retinal image　(b) ground true　(c) Segmented result by the method in []　(d) Segmented result by the method in []　(e) Segmented result by proposed method

图 9　STARE 数据库视网膜血管分割结果比较

Fig. 9　Comparisons of segmentation results on STARE database

噪声的克服能力比较低，对细小的血管分割比较模糊，且(d)中分割出的血管尺寸比标准图像中血管尺寸偏大。这些细小的血管对于分析视网膜疾病有着重要的意义。与以上两种方法相比，本文的方法噪声水平更低，且具有平滑性，分割出的视网膜血管比较清晰，具有更好的鲁棒性和精度。

为了更加突出本文方法分割视网膜血管的优势，我们对 *STARE* 数据集中编号为 im0240 的视网膜图像使用不同方法分割后的结果进行局部放大，如图 10 所示。其中(a)(c)(e)(g)(i)分别为文献[36]、文献[29]、文献[31]、本文方法和专家分割的结果，(b)(d)(f)(h)(j)为与之对应的对右下角部

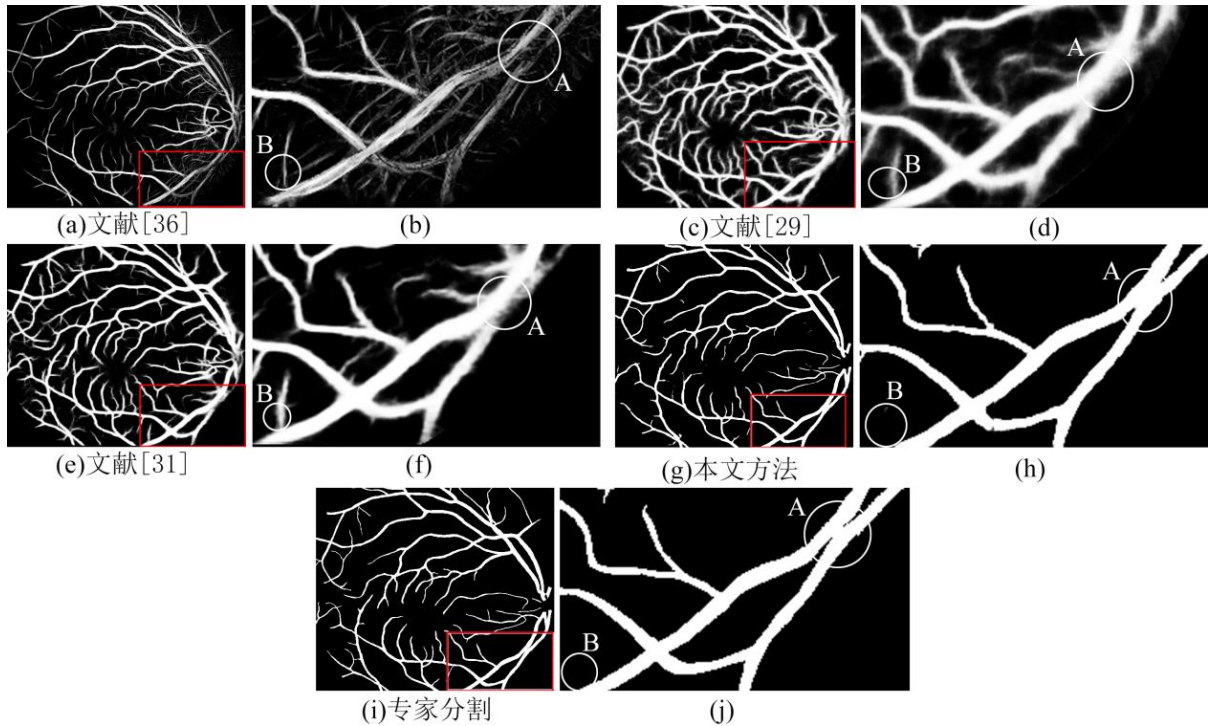

(a)文献[36]　(b)　(c)文献[29]　(d)
(e)文献[31]　(f)　(g)本文方法　(h)
(i)专家分割　(j)

图 10　不同算法的视网膜血管分割局部放大图

Fig. 10 Different methods of partial retinal vessel segmentation

分进行放大后的图像，图中的白色圆圈表示不准确的分割，字母"*A*"和字母"*B*"分别表示邻近的血管和细小的血管。比较(a)(c)(e)(g)可看出，本文的方法分割出的血管更加的清晰，血管的尺寸也比较合适，包含的噪声更少。从(b)中可以看到文献[36]分割出的血管比较虚幻，且不能完整的分割，含有大量的噪声。从 (d)中可以看到文献[29]的分割的血管比实际的血管要大，血管周围出现了大量的噪声使血管比较模糊。 (f)为文献[31] 分割的结果，噪声比前面两种方法要少，分割的结果相对要好，但分割效果还是不理想。在(b)(d)(f)中可以看到其他的方法在 A 处对血管分割不全，无法进行准确的分割，在 B 出都出现的错误的分割。从(h)可以看到，本文方法分割出的血管比较清晰，基本上没有噪声，血管的大小也接近标准大小，含有的错误分割少，分割出的视网膜血管图像更加接近标准的视网膜血管图像。

**6.2 不同分割算法结果的评估**

为了进一步证明本文算法对视网膜血管分割的有效性，在 STARE、DRIVE 数据集上将本文的方法分别与文献[5]、文献[6]、文献[27]、文献[28]、文献[29]、文献[30]、文献[32]、文献[33]、文献[34] 和文献[35]中的方法在敏感性 (Sensitivity)，特效性 (specificity)，准确率 (Accuracy)，F-measure 等几个指标对视网膜血管分割的性能进行了比较。其中灵敏度表示正确分割出的血管像素占真实血管像素的百分比，特异性为正确分割的背景像素占真实背景像素的百

表 2    DRIVE 数据库视网膜血管分割结果
Table 2    Segmentation performance of retinal vessel on the DRIVE database

| Dataset | Methods | Year | F-measure | SE | SP | AC |
|---|---|---|---|---|---|---|
| DRIVE | Chen[27] | 2014 | - | 0.7252 | 0.9798 | 0.9474 |
| | N$^4$-Fields[28] | 2014 | 0.7970 | 0.8437 | 0.9743 | 0.9626 |
| | Azzopardi[5] | 2015 | - | 0.7655 | 0.9704 | 0.9442 |
| | HED[29] | 2015 | 0.6400 | 0.9563 | 0.9007 | 0.9054 |
| | Roychowdhury[30] | 2015 | - | 0.7250 | 0.9830 | 0.9520 |
| | DRIU[31] | 2016 | 0.6701 | 0.9696 | 0.9115 | 0.9165 |
| | Liskowsk[6] | 2016 | - | 0.7763 | 0.9768 | 0.9495 |
| | Qiaoliang Li[32] | 2016 | - | 0.7569 | 0.9816 | 0.9527 |
| | U-Net[33] | 2018 | 0.8142 | 0.7537 | 0.9820 | 0.9531 |
| | Residual U-Net[33] | 2018 | 0.8149 | 0.7726 | 0.9820 | 0.9553 |
| | Recurrent U-Net[33] | 2018 | 0.8155 | 0.7751 | 0.9816 | 0.9556 |
| | R2U-Net[33] | 2018 | 0.8171 | 0.7792 | 0.9813 | 0.9556 |
| | 本文方法 | 2018 | **0.8208** | **0.8274** | **0.9775** | **0.9608** |

表 1    STARE 数据库视网膜血管分割结果
Table 1    Segmentation performance of retinal vessel on the STARE database

| Dataset | Methods | Year | F-measure | SE | SP | AC |
|---|---|---|---|---|---|---|
| STARE | Marin et al.[34] | 2011 | - | 0.6940 | 0.9770 | 0.9520 |
| | Fraz[35] | 2012 | - | 0.7548 | 0.9763 | 0.9534 |
| | HED[29] | 2015 | 0.6990 | 0.5555 | 0.9955 | 0.9378 |
| | Roychowdhury[30] | 2016 | - | 0.7720 | 0.9730 | 0.9510 |
| | Liskowsk[6] | 2016 | - | 0.7867 | 0.9754 | 0.9566 |
| | DRIU[31] | 2016 | 0.7385 | 0.6066 | 0.9956 | 0.9499 |
| | Qiaoliang Li[32] | 2016 | - | 0.7726 | 0.9844 | 0.9628 |
| | U-Net[33] | 2018 | 0.8373 | 0.8270 | 0.9842 | 0.9690 |
| | Residual U-Net[33] | 2018 | 0.8388 | 0.8203 | 0.9856 | 0.9700 |
| | Recurrent U-Net[33] | 2018 | 0.8396 | 0.8108 | 0.9871 | 0.9706 |
| | R2U-Net[33] | 2018 | 0.8475 | 0.8298 | 0.9862 | 0.9712 |
| | 本文方法 | 2018 | **0.8502** | **0.8538** | **0.9878** | **0.9771** |

分比，准确率为正确分割血管和背景像素占整个图像的百分比，F-measure 表示精度和召回率直接的调和均值。

$$Accuracy = \frac{TP+TN}{TP+FN+TN+FP} \quad (7)$$

$$Sensitivity = \frac{TP}{TP+FN} \quad (8)$$

$$specificity = \frac{TN}{TN+FP} \quad (9)$$

$$precision = \frac{TP}{TP+FP} \quad (10)$$

$$recall = \frac{TP}{TP+FN} \quad (11)$$

$$F-measure = 2\times \frac{precison \times sensitivity}{sensitivity+precision} \quad (12)$$

其中，TP 为被正确分割的血管像素的数目，TN 为正确分割的背景像素数目，FP 为错误分割为血管像素的背景像素，FN 为被错误标记为背景像素的血管像素。表 2 和表 1 分别给出了不同方法在 DRIVE 和 STARE 数据集上视网膜分割的准确率。从表 2 和表 1 中可知，本文算法在不同的数据集上敏感度、准确率、F-measure 等都比文献[5]、文献[6]、文献[27]、文献[30]、文献[32]、文献[33]要高，虽然文献[28]、文献[29]、文献[31]的灵敏度比本文方法要高，但是分割出的血管尺寸比实际的要大，且本文算法有最高的 F-measure。对于 DRIVE 数据集上，本文方法对视网膜血管分割的 F-measure 达到 82.08%，比文献[33]高了 0.37%，灵敏度度比文献[33]高了 4.82%。在 STARE 数据集上，我们的方法在 F-measure、灵敏度上比文献[33]分别高了 0.27%、2.4%。因此由表 2 和表 1 中血管分割的评估指标展示可知，在 DRIVE 和 STARE 数据集上，本文的方法在血管分割的各个指标均优于其他视网膜血管分割方法。

图 11 给出了本文方法和其他方法的 F-measure 性能评价曲线，在 DRIVE 数据集上分别与文献[28]、文献[29]、文献[31]、文献[36]进行了比较，在 STARE 数据集上分别与文献[28]、文献[29]、文献[31] 进行了比较。从图 11 可以看出，在 DRIVE 数据集和 STARE 数据集上，上本文方法比文献[28]、文献[29]、文献[31]、文献[36]中的方法效果都要好，且波动比较小，无论在健康还是病变的视网膜图像上，都能进行保

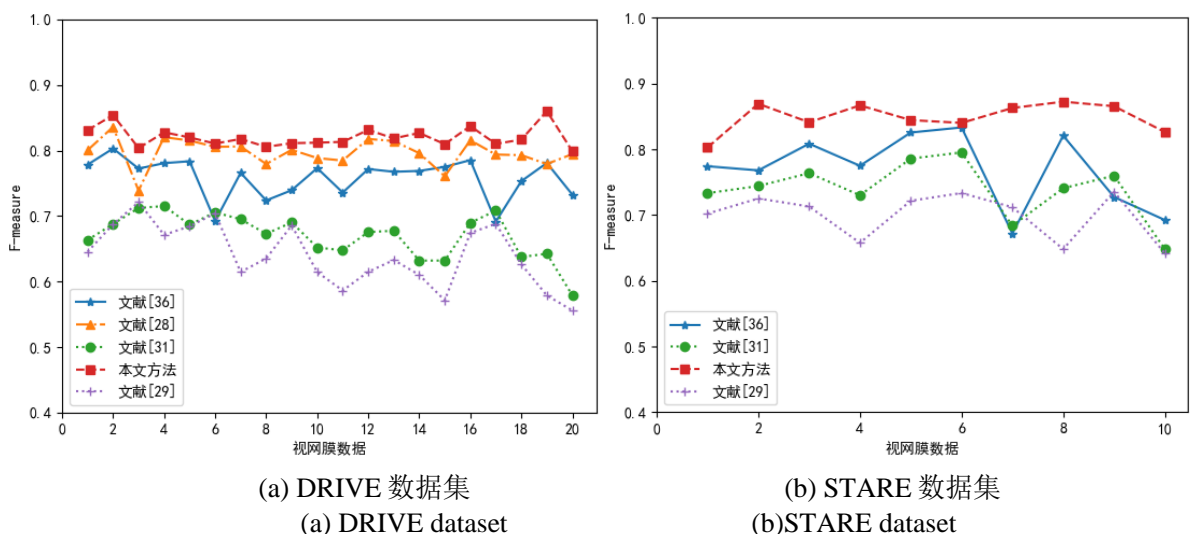

(a) DRIVE 数据集　　　　　　　(b) STARE 数据集
(a) DRIVE dataset　　　　　　　(b)STARE dataset
图 11 不同算法的 F-measure 性能评价曲线
Fig. 11 Different methods of F-measure performance evaluation curve

持相对稳定的分割效果，能够很好的克服病灶的影响。

## 7 结论

视网膜血管的正确分割对帮助医生进行眼底疾病的诊断具有重大的实际意义。本文使用条件深度卷积生成对抗网络对眼底视网膜进行分割，在生成器中使用卷积对图像的特征进行了若干次提取后通过反卷积生成对应的视网膜的血管图像，我们在生成器中加入残差网络模块，由于残差网络对特征值的改变非常敏感性，使得提取的特征更加准确。为了降低模型的规模，在每次进行 3×3 的卷积之前通过 1×1 的卷积进行降维，在保证分割的准确性的情况下减少了参数的数量和计算量。我们分别 DRIVE 和 STARE 数据集上对本文提出的方法的可行性进行了验证，在 DRIVE 和 STARE 数据集上的准确率分别达到了 96.08% 和 97.71%，F-measure 分别达到了 82.08% 和 85.02%，通过对分割出的视网膜血管图像进行分析和比较，本文提出的方法与其他方法相比更具有优势。